# A Quick Review on Recent Trends in 3D Point Cloud Data Compression Techniques and the Challenges of Direct Processing in 3D Compressed Domain

**Mohammed Javed[#], MD Meraz, and Pavan Chakraborty**
Department of IT, Indian Institute of Information Technology, Allahabad, India-211015
[#]Email: javed@iiita.ac.in

Automatic processing of 3D Point Cloud data for object detection, tracking and segmentation is the latest trending research in the field of AI and Data Science, which is specifically aimed at solving different challenges of autonomous driving cars and getting real time performance. However, the amount of data that is being produced in the form of 3D point cloud (with LiDAR) is very huge, due to which the researchers are now on the way inventing new data compression algorithms to handle huge volumes of data thus generated. However, compression on one hand has an advantage in overcoming space requirements, but on the other hand, its processing gets expensive due to the decompression, which indents additional computing resources. Therefore, it would be novel to think of developing algorithms that can operate/analyse directly with the compressed data without involving the stages of decompression and recompression (required as many times, the compressed data needs to be operated or analyzed). This research field is termed as *Compressed Domain Processing*. In this paper, we will quickly review few of the recent state-of-the-art developments in the area of LiDAR generated 3D point cloud data compression, and highlight the future challenges of compressed domain processing of 3D point cloud data.

## 1. Introduction

Due to the global impact of Covid-19, the world is now moving towards digitization and automation at a faster pace than it was anticipated in order to avoid any physical contact with infected humans or contaminated surfaces. Autonomous driving cars (or driverless cars) is a new emerging technology, where a lot of recent efforts can be traced both from academia and Industry [1]. LiDAR (Light Detection And Ranging) sensors are mainly used for navigation of autonomous vehicles as they provide better understanding of objects around them, and also maintain geometric information, in the form of 3D point cloud data. Plenty of applications like segmentation, detection, classification, and tracking are being developed to support this research field [1]. However, 3D point cloud data though preserves significant details of the environment during navigation, but handling in real time is a greater challenge. Therefore researchers have already tried different types of algorithms to reduce the size of 3D point cloud data, by applying different 2D transformations, using graph algorithms, etc. One important classical technique to overcome the huge volume of data is by using different data compression algorithms [1]. However, we know that, if data is captured/stored in the compressed form, it needs to be decompressed first and then processed in order to carry out any type of analytics. Decompression would become very expensive, if real time performance is to be considered. Therefore, the recent trend is to consider processing/operating over the compressed data directly without using the decompression stage is termed as *Compressed Domain Processing* (CDP) [2][3][4]. This provides both computational and storage advantages. CDP has already been successful in many image and video based applications as reported by [2][3][4].

Therefore, the interesting question here is, can we automatically process 3D point cloud compressed directly in 3D Compressed Domain (3DCD)? This review paper aims at reviewing few of the latest 3D point cloud data compression techniques, and providing some pointers to few of the challenges and possibilities of carrying out 3D compressed domain analysis of 3D point cloud data. A pictorial illustration of the conventional compressed domain and 3D point cloud compressed domain is shown in Fig 1.

## 2. Review of recent 3D Point Cloud Data Compression Techniques

Here a quick review of the latest trends in 3D point cloud data compression is provided. 3D point cloud data is obtained from LiDAR devices mounted on autonomous driving vehicles. There are many challenges associated with the LiDAR generated data in order to do any type of processing. However this section is dedicated to discuss the issue of compression 3D LiDAR point cloud data [5-14].

The core idea put forward by researchers in [5] is to employ deep learning driven geometric technique for compressing 3D point cloud raw data, using a hierarchical structure auto-encoder model. The proposed model is novel and has some similarity with PointNet++. The model uses encoder to compress the 3D generated point cloud raw data employing codewords and subsequently further compresses with sparse coding. Exactly the reverse procedure is followed to decompress data with the help of the decoder, generating models at different resolutions. They use Sparse Multiscale Loss Function, and a high compression ratio is achieved outperforming PCL and Draco. The model is tested with ShapeNet40 dataset and has state-of-the-art reconstruction quality.

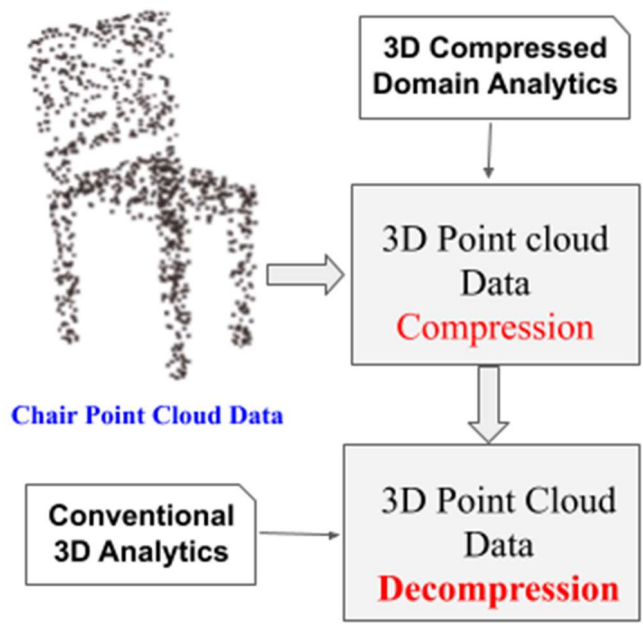

**Fig. 1** *A typical flow diagram showing the conventional 3D Analytics and Compressed Domain 3D Analytics that is anticipated in future*

The researchers in [6], propose the idea of using RNN with Residual Blocks in order to compress 3D point cloud data generated using 3D LiDAR. The method is adaptive to compression ratio and decompression error. Here the raw 3D point cloud data from LiDAR is converted into a 2D matrix, and then pre-processing is done to normalize the data, further RNN is used for compression. This network structure is good for compression as proposed earlier, but for highly accurate decompression they use decoder with residual blocks (not in all layers but only a few for high speed training). The performance metrics used in the paper are bits per point (bpp) to measure size of the data after compression, and Symmetric Nearest Neighbour Root Mean Squared Error to estimate loss incurred after decompression. The good performance measure by the proposed method provides a great potential for its use in real applications like robotics.

A lossy technique for compressing and optimizing 3D point cloud data preserving geometric information is proposed by [7]. They carry out segmentation using region growing technique and subsequently all the points within the enclosed surface are discarded to achieve compression. However, a polynomial equation is used to recover discarded data during decompression. In brief, the 3D point cloud rawa data is divided into different segments and a plane is mapped for each segment. The plane is modelled using a one degree polynomial equation. The method produces a good compression ratio and RMSE when tested with highly structured data. Performance is noted with compression ratio of 89% having RMSE value 0.003 RMSE and within time scale of 0.0643 milliseconds processing time. However the method has the limitation of handling a complex point cloud data.

The survey paper by [8] gives an overview of the existing 3D point cloud compression technology highlighting the design principle, and underscoring their merits and demerits. Different approaches that are proposed in the literature like- 1D traversal, 2D based projection and mapping methods, 3D techniques, octree based methods, LOD, clustering, and transformed representations are discussed. However, 2D techniques are not good for applications like autonomous cars, that require a high accuracy. Therefore it is advised to rely fully on 3D approaches, which offer a better precision through lossy and lossless 3D point cloud compression. It also mentioned that handling of unstructured point cloud data is still an open problem with different challenges. The research paper [9], briefs about basic technologies that are used in 3D Point Cloud Compression. After that it reviews the encoder architectures of TMC1, TMC2 and TMC3 and also TMC13 in detail and then finally analyses their Rate Distortion Performance and their complexities for different cases. It shows TMC2 is the best on average for dense point cloud while TMC13 is optimal for sparse and noisy point cloud with lower time complexity.

This paper [10] is about compressing very dense terrestrial laser scanner (TLS) 3D morphological data generated from trees and forests. Due to inefficient and time consuming previous compression methods, a novel idea based on Compressed Sensing which has brought breakthrough to Shannon-Nyquist sampling theorem. This paper uses the technique of compressed sensing for modelling broad-leaved tree point clouds. For simplifying point clouds and removing outliers Voxel and Statistical Filtering are employed. Then 3D data are divided into three 1D data directly and due to its large length 1D is arranged into individual matrices. Further, sparse transformation is applied and in order to down sample Partial Fourier Matrix is used. For reconstruction of data accurately, ROMP (Regularized Orthogonal Matching Pursuit) is employed. The advantage with this compressed sensing is that it accomplishes compression during the process of sampling, however, in the conventional algorithms, compression is applied after doing full sampling. This method has computational and storage advantages.

Recently tree based structure is used in LiDAR data compression, and here depth of the tree is proportional to resolution of lidar data.In this method First input point cloud is divided into a tree with eight childs, and this process continues till the specified depth. After that tree based Tree-Structured Entropy Model is used to generate the Entropy coding.which further passes it to compact bit represent[11]. There is also an attempt to carry out 3D shape segmentation using deep learning methods [12].

For the first time Chenxi Tu [13] used sequential networks for compressing 3D LiDAR data. Here we simply use the Recurrent neural network method in your work. They divide our complete model in three parts, in the first part they convert the row packet date to a 2D matrix with some additional bits. In part-2 they normalize these data according to sensor specifications. After that in the last part it will pass these data into a RNN based encoder decoder network. On the Decompression side they passed compressed data into a RNN decoder network and after that rearrange these data to get back the original data. Another work by Chenxi [14] for real time compression of streaming 3d point cloud data using Unet based deep learning Network. First they convert raw Lidar point cloud streaming data into 2D matrix form. And then split data into two parts, I-frame and B-fame. After that I-frame will feed into Unet architecture for data interpolation. Then the output of Unet is combined with B-frame for next stage processing. And the last encoder network is used for final point cloud compression.

## 3. Conventional 3D Point Cloud Data Analytics

This section is dedicated to trace the important contributions related to the task of classification, segmentation and tracking of 3D point cloud data that have happened recently. First basic problem in 3D point cloud is classification, for that mainly two types of methods are available. The first one is a projection based method in which the point cloud data is converted into an image based 2D or 3D representation, and then deep learning techniques are applied over them. In the second method a 3D point cloud is directly handled and processed as reported in [1]. But recently many other direct point cloud processing algorithms like Convolution based or Graph based Network get better results.

Second more popular problem in 3D point cloud is object detection [1]. It is considered as one of the major challenges in the self-driving car industry. Here two approaches are usually employed for solving the problem- first approach is region based methods and second one is single shot. The first approach generates possible proposal regions for the object, after that applies classification and bounding box regression algorithms. The second approach is based on two single layer networks that decide object bounding box and class score. This method is faster than method one because it is not a two stage network.

Third common problem in 3D point cloud data is Segmentation. Here the problem is categorised into three categories. First category is Semantic segmentation, second category is Instance segmentation and third category is Part segmentation [1]. In the first category projection based or point based approach is used. In the second problem Instance segmentation proposal based or proposal free method is used. The last one is part Segmentation here [12] a Fully Convolutional Network is used (SFCN), but here the main challenge is 3D shape many folds. So it's difficult to generalize for all parts of the object.

## 4. Future Challenges in Compressed Domain 3D Point Cloud Data Analytics:

The research works presented by [2-4], summarize the various contributions made in the field of image and video processing in the compressed domain focusing on challenges like feature extraction, segmentation, classification, detection, retrieval and so on, all done directly in the compressed data without using any decompression algorithm. Specific to 3D point cloud data that is generated, the basic challenges as underscored in [1] are - carrying out feature extraction, different types of segmentation (semantic, instance, etc), object detection and tracking. Therefore the research work that is to be focused in future will be to do the same above operations directly in the 3D compressed domain of the 3D point cloud data without actually decompression and recompression of data, supporting and achieving real time performance in autonomous driving vehicles. Another important challenge in the direct processing of 3D compressed domain would be applying deep learning models on the compressed data for real time performance which is still an open problem to the researchers around the world. With respect to images and videos applying deep learning models on compressed data in still an unexplored and hot topic of research [2][3].

## 5. Conclusion

This research paper gave a quick review of the latest developments that have happened in the field of 3D point cloud data compression that is being considered as one of the challenges in automatic handling of 3D point cloud LiDAR data generated from autonomous driving vehicles. Further it also highlighted some of the challenges and gave a new perspective for processing 3D point cloud compressed data directly in 3D compressed domain.

*Authors and Affiliations:*
Mohammed Javed[#], MD Meraz, and Pavan Chakraborty
(Department of Information Technology, Indian Institute of Information Technology, Allahabad, Prayagraj, India)
[#]E-mail: javed@iiita.ac.in